\documentclass[review]{elsarticle}

\usepackage{amssymb}
\graphicspath{{./Imgs/}}
\usepackage{color}
\usepackage{amsmath}
\usepackage{bm}
\usepackage{soul}

\usepackage{multirow}
\usepackage{diagbox}
\usepackage{rotating}
\usepackage{booktabs}
\usepackage{colortbl}
\usepackage{overpic}
\usepackage[table]{xcolor}
\usepackage{textcomp}
\usepackage{contour}
\usepackage{textcomp}

\usepackage{subfig}
\usepackage{pdfpages}
\usepackage{lscape}
\usepackage{makecell}

\usepackage{setspace}

\definecolor{mygreen}{RGB}{0,150,0}
\definecolor{myred}{RGB}{200,0,0}
\usepackage[pagebackref=false,letterpaper=true,colorlinks=true,linkcolor=myred,citecolor=blue,bookmarks=false]{hyperref}

\journal{Pattern Recognition}

\newcommand{\figref}[1]{Fig. \ref{#1}}
\newcommand{\tabref}[1]{Table \ref{#1}}

\newcommand{\eqnref}[1]{Eq. \ref{#1}}
\newcommand{\HZeqnref}[1]{(\ref{#1})}

\newcommand{\HZupbf}[1]{\textbf{\textup{#1}}}
\begin{document}

	\begin{frontmatter}
		
		
		
		\title{Contrast-weighted Dictionary Learning Based Saliency Detection for Remote Sensing Images}
		
		\author[uestc]{Zhou Huang}
		\ead{chowhuang23@gmail.com}
		\author[uestc]{Huai-Xin Chen \corref{cor1}}
		\ead{huaixinchen@uesct.edu.cn}
		\author[IIAI]{Tao Zhou}
		\ead{taozhou.ai@gmail.com}
		\author[CETC]{Yun-Zhi Yang}
		\ead{yangyz@cetca.net.cn}
		\author[CETC]{Chang-Yin Wang} 
		\ead{wangcy@cetca.net.cn} 
		\author[uestc]{Bi-Yuan Liu}
		\ead{lby9469@163.com}
		\address[uestc]{University of Electronic Science and Technology of China, Chengdu, China}
		\address[IIAI]{Inception Institute of Artificial Intelligence (IIAI), Abu Dhabi, UAE}
		\address[CETC]{CETC Special Mission Aircraft System Engineering Co.Ltd, Chengdu, China}

		\cortext[cor1]{Corresponding author.}
		
		
		\begin{abstract}
			{\color{myred}Object detection is an important task in remote sensing image analysis. To reduce the computational complexity of redundant information and improve the efficiency of image processing, visual saliency models have been widely applied in this field. In this paper, a novel saliency detection model based on Contrast-weighted Dictionary Learning (CDL) is proposed for remote sensing images. Specifically, the proposed CDL learns salient and non-salient atoms from positive and negative samples to construct a discriminant dictionary, in which a contrast-weighted term is proposed to encourage the contrast-weighted patterns to be present in the learned salient dictionary while discouraging them from being present in the non-salient dictionary. Then, we measure the saliency by combining the coefficients of the sparse representation (SR) and reconstruction errors. Furthermore, by using the proposed joint saliency measure, a variety of saliency maps are generated based on the discriminant dictionary. Finally, a fusion method based on global gradient optimization is proposed to integrate multiple saliency maps. Experimental results on four datasets demonstrate that the proposed model outperforms other state-of-the-art methods.}
			
		\end{abstract}
		
%
		
		\begin{keyword}
			Contrast-weighted dictionary\sep Dictionary learning \sep Gradient optimization \sep Remote sensing\sep Saliency detection
			
		\end{keyword}
		
	\end{frontmatter}
	
	
	\section{Introduction}
	Guided by our gaze, the human visual system (HVS) can quickly and automatically select regions of interest in complex scenes (known as the visual attention mechanism) \cite{Borji2019Salient}. 
	This intelligent mechanism of the HVS has been extensively studied in the fields of psychology \cite{wolfe2004attributes}, neurobiology \cite{mannan2009role}, and computer vision \cite{fan2020taking}. 
	In the past two decades, research on visual saliency has advanced in two ways: eye fixation prediction in human vision \cite{borji2012quantitative} and salient object detection (SOD) in computer vision \cite{tong2015salient,fan2019shifting}. 
	The former focuses on predicting the eye fixations of an observer in a short time \cite{fan2016saliency}, whereas the latter aims to locate or segment the most prominent objects in a scene \cite{cai2019saliency,flores2019saliency,fan2020Camouflage}. 
	Because saliency detection can optimize the computing resources required for image analysis, visual saliency models are widely used in various fields of remote sensing (RS) image processing, including regional change detection \cite{zheng2017unsupervised}, building detection \cite{li2016building} and oil tank detection \cite{liu2019unsupervised}.
	
	The latest research \cite{garcia2012relationship} suggests that information is typically represented by a few simultaneously active neurons. Importantly, while the retina receives a lot information, only a small amount of useful data is transmitted to nerve cells in the visual cortex for processing. This representation of information is known as a sparse representation (SR) \cite{olshausen2004sparse}. 
	The principle of SR is to represent the signal by a linear combination of a series of base vectors in the over-complete dictionary, and that linear combination must be sparse \cite{mairal2009online}. 
	In recent years, image structure analysis based on SR has been widely used in computer vision and image processing. 
	At the same time, SR theory has been introduced into the field of image saliency detection \cite{li2013saliency,zhang2018salient}. 
	However, there are two key problems with SR-based SOD methods: the construction of the SR dictionary and the criteria for saliency measure.
	
	In the construction of dictionaries, most of the early methods used  independent component analysis (ICA) to sample numerous image patches from various kinds of natural images to generate basic atoms \cite{fan2016saliency}. 
	However, these basic atoms cannot create a perfect SR of the detection image without information loss because some features of the training image cannot be accurately captured by the predetermined basic atom. 
	Other SR-based methods \cite{han2011bottom,yan2010visual} usually use the areas around the detected patches for dictionary construction. 
	However, as \cite{shen2012unified} showed, when the salient object has a high contrast with the surrounding patches, such methods usually assign higher values to the edges of the salient object rather than the entire object.
	{\color{myred}In addition, in \cite{li2017multi}, a multi-view joint SR framework that simultaneously considers the inherent contextual structures among instances improved the performance and robustness of the learned dictionary.} 
	Recently, the background prior \cite{wei2012geodesic} was introduced into SR-based saliency detection methods, which assumes that non-salient parts of the image are usually distributed on the boundary. 
	Under this assumption, patches or superpixels near the boundaries of the image are usually selected to build the background dictionary \cite{peng2016salient,xiao2018salient}. 
	However, when the salient object is near the image boundary, some foreground regions are included in the background dictionary, which causes them to be mistakenly detected as background regions. Also, if the background regions near the boundary of the image have distinct features, some background regions will be incorrectly marked as foreground. Moreover, the training sample patches usually have their own characteristic features, such as intensity and contrast, but these are usually disregarded in most existing SOD methods, resulting in salient objects in a scene with similar background and foregrounds being unevenly highlighted.
	
	As for saliency measurement criteria, saliency detection methods based on SR define this in terms of reconstruction error or sparsity of representation coefficients (that is, using the $l_0$-norm to calculate the coding length) \cite{li2013saliency,zhang2018salient,yi2018salient}. 
	These methods also usually add sparse constraints to sparse coefficients to achieve sparse coding of image patches, and they calculate the saliency of image patches by minimizing the sum of the reconstruction errors. Therefore, these representation methods are more sensitive to non-Gaussian noise rather than outliers representing coefficients.
	
	Through our research, we have found that two or more temporary saliency maps are generated in most saliency detection models. 
	Among these methods, some determine the fusion weights through simple weights \cite{zhang2018salient,yi2018salient} or experimental effects \cite{fan2016saliency}. 
	Other methods determine the optimal image fusion weights through methods such as least-squares estimation of training data \cite{xu2016bottom} or Bayesian inference \cite{li2013saliency}, but  do not consider the connections between multiple saliency maps.
	
	{\color{myred}To solve the above problems, we propose an SR method based on Contrast-weighted Dictionary Learning (CDL) for saliency detection. Specifically, this paper uses the positive and negative samples generated by the salient and non-salient regions in the image as a template for dictionary learning.}
	Inspired by the online dictionary learning algorithm \cite{mairal2009online}, to solve the problem of dictionary learning we also propose an online discriminant CDL algorithm, which effectively overcomes the shortcomings of some methods using background priors.
	To determine saliency, we use the $l_2$-norm to measure the sparsity of sparse coefficients, combined with the $l_{1,2}$-norm to calculate the sparse reconstruction errors and improve the expression of “outliers” in the sparse coefficient. For the various saliency maps generated by calculating representation coefficients, we propose an image fusion method based on global gradient optimization to integrate multiple salient images. To summarize, the main contributions of this paper are as follows:
	\begin{itemize}
		\item[(1)] 
		Considering the features of the training sample patch itself, we propose a novel atomic learning formula based on contrast weights. Further, we use an online discriminative CDL to solve the formula.
		\item[(2)] 
		We use the $l_2$-norm to measure the sparsity of sparse coefficients, use the  $l_{2,1}$-norm to measure the sparsity of the reconstruction errors, and then combine the two measures to improve the expression of outliers in the representation coefficients.
		\item[(3)] 
		We use a salient map fusion method based on global gradient optimization to integrate multiple saliency maps. This method optimizes the image fusion effect by establishing the relation between saliency maps.
	\end{itemize}

	The rest of this paper is organized as follows: Section 2 briefly reviews related work. Section 3 describes the SOD method in detail. Sections 4 and 5 give the experimental analysis and conclusions.

	\section{Related Work}
	In recent years, more and more researchers are committed to the work of SOD \cite{zhang2019salient,zhu2018saliency}. Several review papers \cite{Borji2019Salient,fan2018salient} have investigated and discussed many of the most advanced SOD methods in detail. In this section, we review the work most relevant to ours, including SOD based on sparse representations and the application of saliency detection in optical RS images. {\color{myred}Further, SOD based on deep learning is another hot topic in recent years and will be briefly reviewed in this section.}
	\subsection{SOD based on sparse representations}
	In recent years, SR theory has been gradually addressed in the field of saliency detection. Generally, SR based saliency detection methods need to first construct an over-complete dictionary, then sparsely represent an input image through the dictionary, and finally measure the saliency according to the SR coefficients or reconstruction errors. 
	In \cite{fan2016saliency}, the construction of the dictionary was learned by applying ICA on the image patches sampled from each position of the input image and using the reconstruction errors to measure the saliency. 
	In the method of \cite{han2011bottom} the image patches around the central patch were used for SR, and the saliency was measured by the coding length or residual. These methods usually give higher saliency values to the object boundaries, because both the background and foreground are included in the dictionary. 
	
	Later, the background prior method \cite{wei2012geodesic} was proposed. As an extension of this, some methods then \cite{peng2016salient,xiao2018salient} used patches or superpixels near the image boundary as background templates to construct a global background dictionary and sparsely reconstruct the image. Recently, in \cite{liu2018salient}, a SOD method was proposed based on two-stage graphs, taking into account the consistency of adjacent spaces between graph nodes and the consistency of regional spaces, while improving the accuracy of SOD in complex scenes.
	\subsection{Application of SOD in RS images}
	Due to the rapid development of massive RS data and the complexity of RS scenes, many traditional methods of processing natural images are not suitable for RS images. As one method of data compression and rapid screening, saliency detection can effectively process RS data. Importantly, there are several essential similarities between SOD/ target detection and extraction in RS images. For instance, both extract regions of interest in an image based on the saliency of a particular task or target. As image processing and RS technology have developed, saliency detection has been widely used in the field of RS. Many researchers have combined visual saliency and image interpretation to accomplish specific target detection, such as regional change detection \cite{zheng2017unsupervised}, airport detection \cite{yao2015coarse}, building detection \cite{li2016building} and oil tank detection \cite{liu2019unsupervised}. For example, Yao et al. \cite{yao2015coarse} proposed a coarse-to-fine airport saliency detection model. At the coarse layer, combined with contrast and linear density clues, a goal-oriented saliency model was established to quickly locate airport candidate regions. 
	Later, Li \cite{li2016building} et al. proposed a two-step building extraction method based on saliency cues, designed a saliency estimation algorithm for building objects, extracted saliency cues in a local region of each candidate building, and integrated them into a probability model to get the final building extraction results. However, these methods do not involve road detection methods based on saliency in RS images.
	\subsection{SOD based on deep learning}
	{\color{myred}Recently, SOD methods based on deep learning have attracted more attention. Zhang et al. \cite{zhang2018bi} proposed a SOD model based on fully a convolutional neural network by introducing a gated two-way message passing module to integrate multi-level features. In \cite{qin2019basnet}, a predict-refine architecture and a new hybrid loss for boundary-aware SOD were proposed to pay more attention to the boundary quality of salient objects. In another work focusing on salient edge information \cite{wang2019salient}, an essential pyramid attention structure for SOD was designed to enhance the representation ability of the corresponding network layer. }
	
	{\color{myred}In order to prevent SOD methods for RGB images from failing \cite{wei2012geodesic,fan2018salient,Fu2020JLDCF} when processing complex scenes, as is common in recent saliency detection works dedicated to RGB-D, \cite{Zhang2020UCNet} introduced the probabilistic RGB-D saliency detection network via conditional variational autoencoders to model human annotation uncertainty and generate multiple saliency maps for each input image by sampling in the latent space. Further, Fan et al. \cite{fan2020rethinking} constructed a 1K high-resolution saliency person dataset, and proposed a baseline architecture called the Deep Depth-Depurator Network for saliency detection. In addition, binocular stereo cameras are widely used in various tasks of RS photogrammetry, which makes it possible to use supplementary depth information to further accurately detect and identify targets in the field of RS.}
	\section{Proposed Saliency Detection Model}
	This section describes the proposed saliency detection model in detail. As shown in~\figref{fig:FlowChart}, the model includes three main parts: CDL-based discriminant dictionary learning, saliency maps generation and fusion.
	\begin{figure}[thp!]
		\centering
		\begin{overpic}[width=.91\columnwidth]{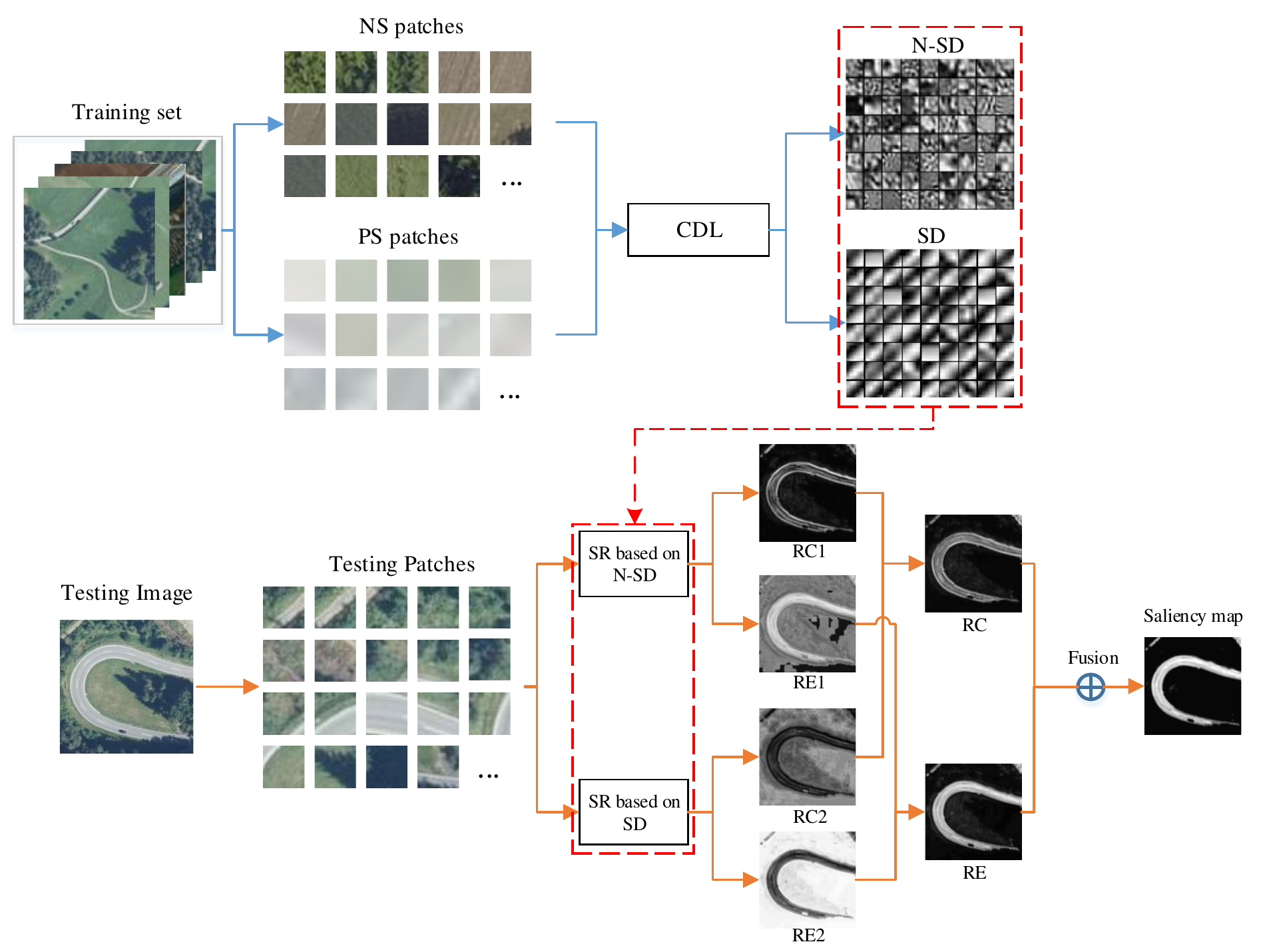}
		\end{overpic}
		\caption{
			 Diagram of the proposed SOD method. NS: negative sample, PS: positive sample, CDL: contrast-weighted dictionary learning, N-SD: non-salient dictionary, SD: salient dictionary, SR: sparse representation, RC: representation coefficient, RE: reconstruction error.
		}\label{fig:LCWAFlowChart}
	\end{figure}
	\subsection{Contrast-weighted dictionary learning formula}
	In the image processing method based on SR, an image patch is usually represented by a linear combination of a few atoms in an over-complete dictionary $\HZupbf{D} =\left\{\HZupbf{d}_i\right\}_{i=1}^k\in\mathbb{R}^{n\times k}  $; that is, the image patch $ \HZupbf{x}\in\mathbb{R}^n $ is estimated by dictionary $ \HZupbf{D} $ and the calculated sparse coefficients $ \bm{\alpha}\in\mathbb{R}^k $ . The equation is
	\begin{equation}\label{equ:1}
	\HZupbf{x}=\HZupbf{D}\bm{\alpha}\quad s.t.\quad \left\|\HZupbf{x}-\HZupbf{D}\bm{\alpha}\right\|_2\leq \xi,
	\end{equation}
	where $ \left\|\bullet\right\|_2 $ is the $ l_2 $-norm used to measure the deviation and $ \xi $ is the error. Within the feasible set, the solution that minimizes the number of nonzero sparse coefficients is undoubtedly an attractive representation. This form can be expressed as:
	\begin{equation}\label{equ:2}
	\min\limits_{\bm{\alpha}\in\mathbb{R}^k}\left\|\HZupbf{x}-\HZupbf{D}\bm{\alpha}\right\|_{2}^{2}\quad s.t.\quad \left\|\bm{\alpha}\right\|_0\leq L,
	\end{equation}
	where $ L $ is the sparsity of the coefficients $ \bm{\alpha} $ . In \eqnref{equ:2}, the atoms in $ \HZupbf{D} $ represent the smallest unit in the reconstructed image patches. Here, the atoms in $ \HZupbf{D} $ need to be learned from the training patches $ \HZupbf{X}=\left\{ \HZupbf{x} \right\}_{i=1}^{m} $, which can be achieved by [19]
	\begin{equation}\label{equ:3}
	\min\limits_{\HZupbf{D,A}}\frac{1}{m} \sum_{i=1}^{m}\left( \frac{1}{2} \left\|\HZupbf{x}_{i}-\HZupbf{D}\bm{\alpha}_i \right\|_{2}^{2} +\lambda \left\|\bm{\alpha}_i \right\|_1 \right),
	\end{equation}
	where $ \lambda $ is the trade-off between the reconstruction errors $ \left\|\HZupbf{x}_{i}-\HZupbf{D}\bm{\alpha}_i \right\|_{2}^{2} $ and the sparsity of the coefficient $ \left\|\bm{\alpha}_i \right\|_1 $ and $\HZupbf{A} =\left\{\bm{\alpha _i}\right\}_{i=1}^{m} $ is the SR coefficients set corresponding to $ \HZupbf{X} $ . According to \eqnref{equ:3}, we study salient and non-salient dictionary learning based on contrast-weighted atoms. As one of the features of an image, contrast plays an important role in both local and global saliency detection methods \cite{zhao2019contrast,cheng2014global}. 
	{\color{myred} In order to enhance the ability of the base atom to learn the image contrast features and improve the sensitivity to the contrast of the surrounding pixels, a novel contrast-weighted term is incorporated in our formulation to encourage/discourage the contrast-weighted patterns in the learned salient/non-salient dictionary, respectively. More specifically, in our weight function, the weight of each pixel in the base atom is calculated by the relative brightness contrast of the pixels and the corresponding training sample patch $ \HZupbf{x}_{i} $. Thus, the weight of the $ j $-th pixel $ p_{ij} $ in the $ i $-th sample patch is:
	\begin{equation}\label{equ:4}
	\textbf{\textup{w}}\left( {p}_{ij}\right)=\frac{Lum\left( {p}_{ij} \right)-\textup{mean}\left(\emph{Lum}\left( \HZupbf{x}_{i} \right) \right)}{\textup{max}\left(\emph{Lum}\left( \HZupbf{x}_{i} \right) \right)},
	\end{equation}
	where $ Lum\left( \cdot \right) $ is the luminance value operator for calculating the sample patch, and $ \textup{mean}\left( \cdot \right) $ and $ \textup{max}\left( \cdot \right) $ are the average value operator and the maximum value operator, respectively. 
	Note that in the actual calculation, the $ i $-th sample patch is treated as a column vector; that is, $ \textbf{\textup{w}}\left( {p}_{ij}\right) $ can be expressed as $ \textbf{\textup{w}}_{ij}^{T}\in\mathbb{R}^{1 \times n} $ , and $ n $ is the number of pixels of the sample patch. 
	Upon $ \HZupbf{W}_{i}^T\in\left\{ \textbf{\textup{w}}_{ij}^{T}\right\}_{i,j=1}^{m,n} $, the contrast weight term can be designed by $ \left\|\textbf{\textup{W}}_{i}^{T}\HZupbf{D} \right\|_{2}^{2} $, which quantifies the degree of weighted contrast. Thus, given the contrast weight term, by rewriting~\eqnref{equ:3}, we have the following formula for optimizing the salient and non-salient dictionary learning:}
	\begin{equation}\label{equ:5}
	\min\limits_{\HZupbf{D}^\textup{H},\HZupbf{A}^\textup{H}}\frac{1}{m^\textup{H}} \sum_{i=1}^{m^\textup{H}}\left( \frac{1}{2} \left\|\HZupbf{x}_{i}^\textup{H}-\HZupbf{D}^\textup{H}\bm{\alpha}_{i}^\textup{H} \right\|_{2}^{2}+\lambda_1\left\| \bm{\alpha}_{i}^\textup{H} \right\|_{1} + \lambda_2\left\|\textbf{\textup{W}}_{i}^{T}\HZupbf{D}^\textup{H} \right\|_{2}^{2} \right),\textup{H}=\left\{\ \textup{P} \vee\textup{N} \right\},
	\end{equation}
	where $ m^\textup{H} $ represents the number of positive or negative samples, $ \left\{\bm{\alpha}_{i}^\textup{H}  \right\}_{i=1}^{m^\textup{H}} $ is the SR coefficient of positive or negative sample patches, and $ \HZupbf{D}^\textup{H} $ is a salient or non-salient dictionary trained from positive or negative samples. 
	The meaning of $ \lambda_1 $ is the same as that of $ \lambda $  in formula  \HZeqnref{equ:3}, and $ \lambda_2 $ (a very small positive number) is a regularization parameter that controls the influence of contrast-weighted terms. 
	\subsection{The solution to the dictionary learning formulation}
	We can learn the salient and non-salient dictionaries through \HZeqnref{equ:5}. Because the online dictionary learning algorithm \cite{mairal2009online} can deal with large dynamic datasets and is faster than the batch algorithm, we propose the CDL algorithm to solve \HZeqnref{equ:5}. Similar to the standard dictionary learning algorithm, we divide the optimization problem in \HZeqnref{equ:5} into two subprocesses to solve alternately; namely, the SR and dictionary update. Specifically, the initialization training dictionary $ \HZupbf{D}^\textup{H} $ is generally obtained by randomly sampling the training sample set. Thus, the first step is to fix $ \HZupbf{D}^\textup{H} $ , and the sparse coefficient $ \HZupbf{A}^\textup{H}= \left\{\bm{\alpha}_{i}^\textup{H}  \right\}_{i=1}^{m^\textup{H}} $ can then be obtained by the SR method. 
	The second step is to fix $ \HZupbf{A}^\textup{H} $ , and the updated dictionary $ \HZupbf{D}^\textup{H} $ can then be solved by the dictionary update method. The first and second steps of the iteration are done until convergence is reached.
	\subsubsection{Sparse representation}
	From the above, it can be seen that the solution to \HZeqnref{equ:5} is an iterative optimization process, assuming that in the $ i $-th iteration, $ \HZupbf{x}_{t}^{\textup{H}} $ is a randomly selected image patch from the training set, and $ \bm{\alpha}_{t}^{\textup{H}} $ is the coefficient of  $ \HZupbf{x}_{t}^{\textup{H}} $ obtained by the $ \left(t-1\right) $-th updated dictionary $ \HZupbf{D}_{t-1}^{\textup{H}} $ through the SR algorithm. 
	Because the contrast-weighted term $ \lambda_2\left\|\textbf{\textup{W}}_{i}^{T}\HZupbf{D}^\textup{H} \right\|_{2}^{2} $ in \HZeqnref{equ:5} is independent of the sparse coefficient $ \bm{\alpha}_{i}^{\textup{H}} $ , the sparse coefficient a in the $ i $-th iteration can be expressed as
	\begin{equation}\label{equ:6}
	 \alpha_{t}^{\textup{H}}\triangleq\mathop{\arg\min}_{\alpha_{t}^{\textup{H}}\in\mathbb{R}^{k}}\frac{1}{2} \left\|\HZupbf{x}_{t}^\textup{H}-\HZupbf{D}_{t-1}^\textup{H}\bm{\alpha}_{t}^\textup{H} \right\|_{2}^{2}+\lambda_1\left\| \bm{\alpha}_{t}^\textup{H} \right\|_{1},
	\end{equation}
	
	The SR problem of the above fixed dictionary is the $ l_1 $-regularized linear least square problem. In this paper, the LARS-Lasso algorithm \cite{osborne2000new} is used to solve this problem.
	\subsubsection{Dictionary update}
	After the SR step of the $ t $-th iteration, the sparse coefficient $ \left\{ \alpha _{i}^{\textup{H}}\right\}_{i=1}^{t} $ of the image patch $ \left\{ \HZupbf{x} _{i}^{\textup{H}}\right\}_{i=1}^{t} $ after training is obtained. In the $ t $-th iteration, with fixed $ \alpha _{i}^{\textup{H}} $, the dictionary can be updated using the following optimization function according to \HZeqnref{equ:5}:
	\begin{equation}\label{equ:7}
	\HZupbf{D}_{t}^{\textup{H}}\triangleq\mathop{\arg\min}_{\HZupbf{D}_{t}^{\textup{H}}\in\mathbb{R}^{n \times k}}  \frac{1}{t} \sum_{i=1}^{t} \left( \frac{1}{2} \left\|\HZupbf{x}_{i}^\textup{H}-\HZupbf{D}_{t-1}^\textup{H}\bm{\alpha}_{t}^\textup{H} \right\|_{2}^{2}+\lambda_1\left\| \bm{\alpha}_{t}^\textup{H} \right\|_{1} + \lambda_2\left\|\textbf{\textup{W}}^{T}\HZupbf{D}_{t-1}^{\textup{H}} \right\|_{2}^{2} \right),
	\end{equation}
	where $ \HZupbf{D}_{t}^{\textup{H}} $ is the discriminant dictionary obtained after the $ t $-th iterative learning. 
	
	Because the patch coordinate descent algorithm \cite{wright2015coordinate} has the advantages of no parameters and no need for any learning rate adjustment, we updated each atom of the dictionary using this algorithm. For example, the $ j $-th atom $ \HZupbf{d}_{j,t}^{\textup{H}} $ for updating the dictionary in the $ t $-th iteration is calculated by
	\begin{equation}\label{equ:8}
	\begin{aligned}
	 \HZupbf{d}_{j,t}^{\textup{H}}=\HZupbf{d}_{j,t-1}^{\textup{H}}-\frac{\sigma}{t}\frac{\partial}{\partial \HZupbf{d}_{j}^{\textup{H}}} \left[ \sum_{i=1}^{t} \left( \frac{1}{2} \left\|\HZupbf{x}_{i}^\textup{H}-\widehat{\HZupbf{D}}_{j,t}^\textup{H}\bm{\alpha}_{i}^\textup{H} \right\|_{2}^{2}+\lambda_1\left\| \bm{\alpha}_{i}^\textup{H} \right\|_{1} 
	 \right. \right. 
	 \\
	  \left.\left.
	  +\lambda_2\left\|\textbf{\textup{W}}_{j}^{T}\widehat{\HZupbf{D}}_{j,t}^{\textup{H}} \right\|_{2}^{2} \right) \right]_{|\mathrm{d_{j,t-1}^{\textup{H}}}}.
	\end{aligned}	
	\end{equation}
	
	For convenience, let
	\begin{equation}\label{equ:9}
	\textbf{M}=  \sum_{i=1}^{t} \left( \frac{1}{2} \left\|\HZupbf{x}_{i}^\textup{H}-\widehat{\HZupbf{D}}_{j,t}^\textup{H}\bm{\alpha}_{i}^\textup{H} \right\|_{2}^{2}+\lambda_1\left\| \bm{\alpha}_{i}^\textup{H} \right\|_{1} 
	+\lambda_2\left\|\textbf{\textup{W}}_{j}^{T}\widehat{\HZupbf{D}}_{j,t}^{\textup{H}} \right\|_{2}^{2} \right).
	\end{equation}
	
	In \eqnref{equ:8}, $ \sigma $ is the learning rate of gradient descent, and $  \widehat{\HZupbf{D}}_{j,t}^{\textup{H}}=[\HZupbf{d}_{1,t}^{\textup{H}},\HZupbf{d}_{2,t}^{\textup{H}},\cdots , $ $\HZupbf{d}_{j,t}^{\textup{H}},\HZupbf{d}_{j+1,t-1}^{\textup{H}},\cdots ,\HZupbf{d}_{k,t-1}^{\textup{H}} ]$, Note that the only variable that must be updated is $ \HZupbf{d}_{j,t}^{\textup{H}} $ in $ \widehat{\HZupbf{D}}_{j,t}^{\textup{H}} $. 
	After \eqnref{equ:8} for the current iteration, the previous $ j $ atoms, that is, $ \left\{\HZupbf{d}_{1,t}^{\textup{H}},\HZupbf{d}_{2,t}^{\textup{H}},\cdots ,\HZupbf{d}_{j,t}^{\textup{H}} \right\} $, are updated. Using the trace $ Tr\left( \bullet \right) $ of the matrix to represent the $ l_2 $-norm and then expressing it as the derivative of $ \HZupbf{d}_{j}^{\textup{H}} $ , \eqnref{equ:9} can be rewritten as
	\begin{equation}\label{equ:10}
	\begin{aligned}
	\frac{\partial}{\partial \HZupbf{d}_{j}^{\textup{H}}}\left( \textbf{M} \right)_{|\mathrm{d_{j,t-1}^{\textup{H}}}}=\frac{1}{2}\frac{\partial}{\partial \HZupbf{d}_{j}^{\textup{H}}}Tr\left[ \left( \widehat{\HZupbf{D}}_{j,t}^{\textup{H}} \right)^{T}\widehat{\HZupbf{D}}_{j,t}^{\textup{H}}\HZupbf{B}_{t}^{\textup{H}}\right]-
	\frac{\partial}{\partial \HZupbf{d}_{j}^{\textup{H}}}Tr\left[ \left( \widehat{\HZupbf{D}}_{j,t}^{\textup{H}} \right)^{T}\HZupbf{C}_{t}^{\textup{H}}\right]
	\\
	+\frac{\partial}{\partial \HZupbf{d}_{j}^{\textup{H}}}Tr\left[ \lambda_{2}t\HZupbf{W}_{j}^{T} \widehat{\HZupbf{D}}_{j,t}^{\textup{H}} \left(
	 \widehat{\HZupbf{D}}_{j,t}^{\textup{H}} \right)^{T}\HZupbf{W}_{j}\right],
	\end{aligned}
	\end{equation}
	where $ \HZupbf{B}_{t}^{\textup{H}} $ and $ \HZupbf{C}_{t}^{\textup{H}} $ are defined as $ \sum_{i=1}^{t}\alpha_{i}^{\textup{H}}\left(\alpha_{i}^{\textup{H}}\right)^{T} $ and $ \sum_{i=1}^{t}\HZupbf{x}^{\textup{H}}\left(\alpha_{i}^{\textup{H}}\right)^{T} $ , which refer to storing all the information of the sparse coefficients and sparsely represented image patches of all previous iterations, respectively. According to the derivative calculation rule of the matrix trace, \eqnref{equ:10} can be expressed as
	\begin{equation}\label{equ:11}
		\frac{\partial}{\partial \HZupbf{d}_{j}^{\textup{H}}}\left( \textbf{M} \right)_{|\mathrm{d_{j,t-1}^{\textup{H}}}}=\widehat{\HZupbf{D}}_{j,t}^{\textup{H}}\HZupbf{b}_{j,t}^{\textup{H}}-\HZupbf{c}_{j,t}^{\textup{H}}
		+2\lambda_{2}t\HZupbf{W}_{j}\HZupbf{W}_{j}^{T}\HZupbf{d}_{j}^{\textup{H}},
	\end{equation}
	where $ \HZupbf{b}_{j,t}^{\textup{H}} $ and $ \HZupbf{c}_{j,t}^{\textup{H}} $ represent the $ j $-th columns of $ \HZupbf{B}_{j,t}^{\textup{H}} $ and $ \HZupbf{C}_{j,t}^{\textup{H}} $ respectively. Thus \eqnref{equ:8} can be rewritten as
	\begin{equation}\label{equ:12}
	\HZupbf{d}_{j,t}^{\textup{H}}=\HZupbf{d}_{j,t-1}^{\textup{H}}-\frac{\sigma}{t}\left( \widehat{\HZupbf{D}}_{j,t}^{\textup{H}}\HZupbf{b}_{j,t}^{\textup{H}}-\HZupbf{c}_{j,t}^{\textup{H}} \right)-2\lambda_{2}\sigma\HZupbf{W}_{j}\HZupbf{W}_{j}^{T}\HZupbf{d}_{j,t-1}^{\textup{H}}.
	\end{equation}
	
	According to \cite{mairal2009online}, the $ \sigma/t $ in \eqnref{equ:12} can be expressed approximately as $ 1/\HZupbf{B}_{j}^{\textup{H}}\left(j,j\right) $ . When all the atoms $ \left\{ \HZupbf{d}_{j,t}^{\textup{H}} \right\}_{j=1}^{k} $   are updated, the dictionary $ \HZupbf{D}_{t}^{\textup{H}} $   completes the $ t $-th learning.
	
	In summary, after the iterative SR and dictionary update steps, we obtain salient and non-salient dictionaries. \tabref{tab:1} summarizes our CDL algorithm.
	
\begin{table}[t!]
	\centering
	\scriptsize
	\caption{Summary of the CDL algorithm}\label{tab:1}
	\begin{tabular*}{0.98\linewidth}{l}
		\toprule 
		\textbf{Algorithm 1}. Online discriminant dictionary learning algorithm based on weighted contrast \\
		\midrule
		~\textbf{Input:} Vectorised training patches $ \HZupbf{X}^{\textup{H}}\in\mathbb{R}^{n\times m} $. \\
		~\textbf{Output:} The learned dictionary $ \HZupbf{D}^{\textup{H}}\in\mathbb{R}^{n\times k} $. \\
		~\textbf{Initialization:} The contrast-weighted matrix $ \HZupbf{W}^{T} $ is obtained by \eqnref{equ:4}; \\
		~~~~~~~~~~~~~~~~~~~~~~~Randomly select the samples in the training set to fill $ \HZupbf{D}_{0}^{\textup{H}} $;  \\
		~~~~~~~~~~~~~~~~~~~~~~~Set $ \HZupbf{B}_{0}^{\textup{H}}\in\mathbb{R}^{k\times k} $ and $ \HZupbf{C}_{0}^{\textup{H}}\in\mathbb{R}^{n\times k} $ to zero matrices;  \\
		~~~~~~~~~~~~~~~~~~~~~~~Regularization parameter $ \lambda_1 $ and $ \lambda_2 $ ;\\
		~~~~~~~~~~~~~~~~~~~~~~~Number of iterations $ T $.\\
		1. \textbf{For} $ t=1 $ to  $ T $ \textbf{do}\\
		2.~~~~~~~~Randomly select the image patches $ \HZupbf{X}^{\textup{H}} $ from the training set $ \HZupbf{x}_{t}^{\textup{H}}\in\mathbb{R}^{k\times 1} $.  \\
		3.~~~~~~~~Sparse coding:\\
		~~~~~~~~~~~~~~~~~~~~~~~Obtained $ \alpha_{t}^{\textup{H}}\in\mathbb{R}^{k\times 1} $ by solving \eqnref{equ:6} with LARS-Lasso [60] algorithm.\\
		4.~~~~~~~~Update $ \HZupbf{B}_{t}^{\textbf{H}} $ and $ \HZupbf{C}_{t}^{\textbf{H}} $: \\
		~~~~~~~~~~~~~~~~~~~~~~~$ \HZupbf{B}_{t}^{\textbf{H}}=\sum_{i=1}^{t}\alpha_{i}^{\textup{H}}\left(\alpha_{i}^{\textup{H}}\right)^{T}=\HZupbf{B}_{t-1}^{\textbf{H}}+\alpha_{t}^{\textup{H}}\left(\alpha_{t}^{\textup{H}}\right)^{T}  $,\\
		~~~~~~~~~~~~~~~~~~~~~~~$ \HZupbf{C}_{t}^{\textbf{H}}=\sum_{i=1}^{t}\HZupbf{x}_{i}^{\textup{H}}\left(\alpha_{i}^{\textup{H}}\right)^{T}=\HZupbf{C}_{t-1}^{\textbf{H}}+\HZupbf{x}^{\textup{H}}\left(\alpha_{t}^{\textup{H}}\right)^{T}  $.\\
		5.~~~~~~~~Dictionary update:\\
		~~~~~~~~~~~~~~~~~~~~~~~\textbf{For} $ t=1 $ to  $ T $ \textbf{do}\\
		~~~~~~~~~~~~~~~~~~~~~~~$\HZupbf{d}_{j,t}^{\textup{H}}=\HZupbf{d}_{j,t-1}^{\textup{H}}-\frac{1}{\HZupbf{B}_{j}^{\textup{H}}\left(j,j\right)} \left(  \widehat{\HZupbf{D}}_{j,t}^{\textup{H}}\HZupbf{b}_{j,t}^{\textup{H}}-\HZupbf{c}_{j,t}^{\textup{H}} \right)-2\lambda_{2}\sigma\HZupbf{W}_{j}\HZupbf{W}_{j}^{T}\HZupbf{d}_{j,t-1}^{\textup{H}} $,\\
		~~~~~~~~~~~~~~~~~~~~~~~\textbf{End For}\\
		6.~~~~~~~~Obtain the discriminant dictionary $ \HZupbf{D}_{t}^{\textup{H}}=\left[\HZupbf{d}_{1,t}^{\textup{H}},\HZupbf{d}_{2,t}^{\textup{H}},\cdots ,\HZupbf{d}_{k,t}^{\textup{H}} \right] $ for the current iteration. \\
		7.\textbf{End For}\\
		8.Return:The learned dictionary $ \HZupbf{D}^{\textup{H}}=\HZupbf{D}_{T}^{\textup{H}} $.\\
		\bottomrule
	\end{tabular*}
\end{table}
\subsection{Saliency image generation}
This subsection describes the saliency measurement criteria based on SR coefficients and reconstruction errors.
\subsubsection{Saliency measure based on sparse representation coefficients}
\begin{figure}
	\centering
	\subfloat[]{\includegraphics[scale=0.58]{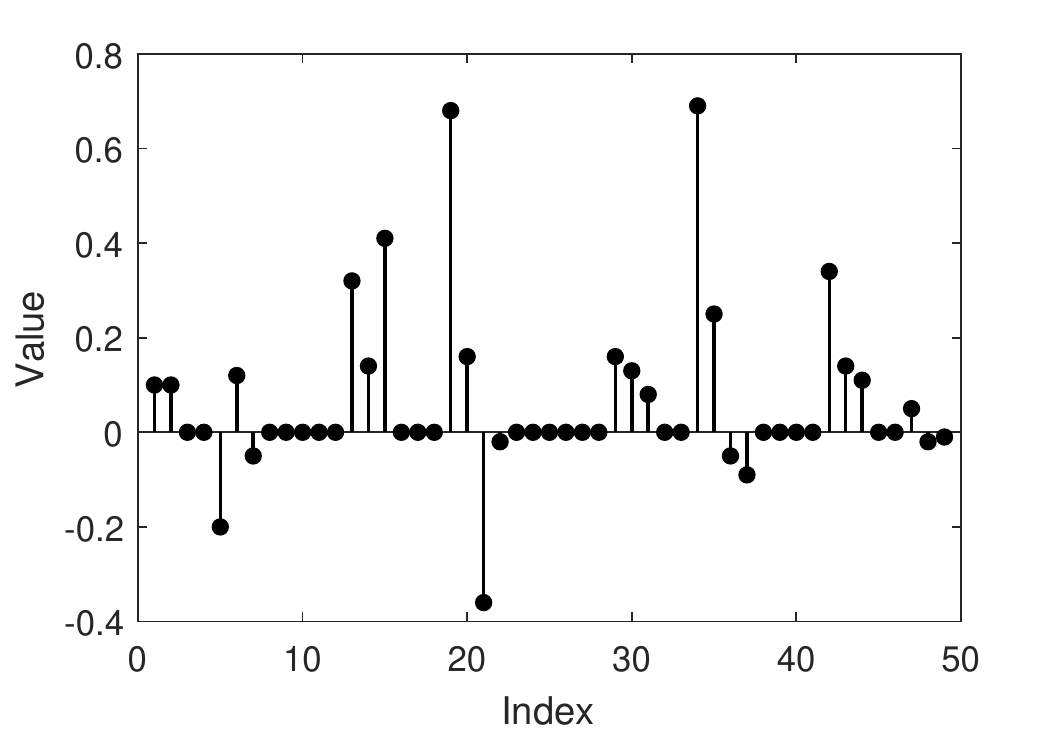}}%
	\subfloat[]{\includegraphics[scale=0.58]{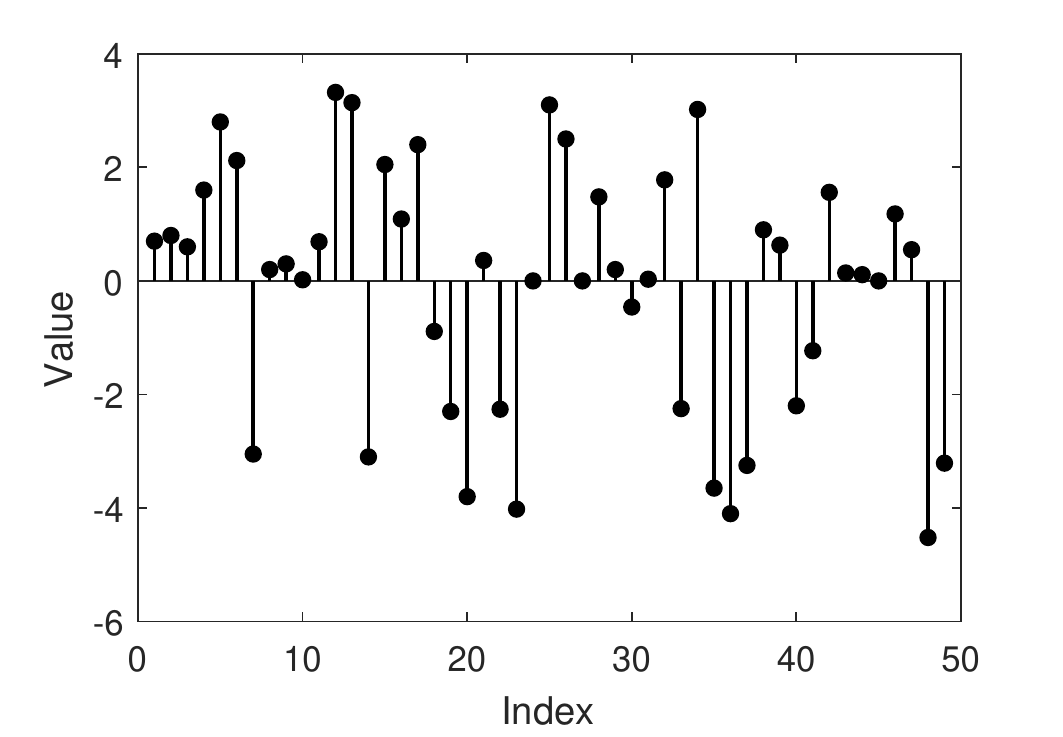}}%
	\caption{Comparison of SR coefficients using salient dictionary. (a) SR coefficients of salient patches. (b) SR coefficients of non-salient patches.}%
    \label{fig:Outliers}
\end{figure}

{\color{myred}In the saliency detection process, the saliency of each pixel can be measured to a certain extent by the representation coefficient of an image patch centered on the pixel, where the different representation coefficients of the image patch $ \HZupbf{x}_{i} $ are calculated by the discrimination dictionary $ \HZupbf{D}^{H} $ through the following formula:
\begin{equation}\label{equ:013}
\alpha_{i}^{H}=\mathop{\arg\min}\frac{1}{2}\left\|\HZupbf{x}_{i}-\HZupbf{D}^{H} \alpha_{i}^{H} \right\|_{2}^{2}+\lambda_1\left\| \alpha_{i}^{H} \right\|_{1},
\end{equation}
where $ \lambda_1 $ has same meaning as $ \lambda $ in \eqnref{equ:3}. As shown in \figref{fig:Outliers}, when using the salient dictionary for sparse reconstruction, non-salient image patches obtain their SR coefficients with high energy, while salient image patches obtain their SR coefficients with lower energy. This is because the salient dictionary has high contrast with non-salient image patches, and the saliency image patches have low contrast. On the basis of this observation, we define the saliency measure of a pixel as:
\begin{equation}\label{equ:13}
S_{\HZupbf{A}\left( i \right)}=1-\exp\left( -\frac{\left\|\alpha_{i}^{N} \right\|_{2}^{2}-\left\|\alpha_{i}^{P} \right\|_{2}^{2}}{2\eta_{\HZupbf{A}}^{2}} \right),
\end{equation}
where $ \alpha_{i}^{N} $ and $ \alpha_{i}^{P} $ represent the representation coefficients obtained by \eqnref{equ:013} for the image patch centered on pixel $ i $, and $ \eta_{\HZupbf{A}}  $ is a scalar parameter, which is set to 1 in the experiment.} 
\subsubsection{Saliency measurement based on reconstruction error}
Reconstruction error is widely used in saliency detection based on SR. Generally speaking, an image patch has a larger relative reconstruction error for the discriminant dictionary, so it will have a greater saliency value. Therefore, we define the saliency measure of pixels based on SR coefficients as:
\begin{equation}\label{equ:14}
S_{\HZupbf{R}\left( i \right)}=1-\exp\left( -\frac{\min\limits_{\alpha_{i}^{N}}\left\|x_i-D^N\alpha_{i}^{N} \right\|_{2,1}-\min\limits_{\alpha_{i}^{P}}\left\|x_i-D^P\alpha_{i}^{P} \right\|_{2,1}}{2\eta_{\HZupbf{R}}^{2}} \right),
\end{equation}
where $ x_i $ is the image patch centered on a pixel $ i $ , $ D^N  $ and $ D^P $ represent the non-salient and salient dictionary, respectively, $ \alpha_{i}^{N} $ and $ \alpha_{i}^{P} $ are the representation coefficients obtained by the discriminant dictionary, and $ \eta_{\HZupbf{R}} $ is the scale parameter and is set to 1 in the experiment.
\subsection{Saliency map fusion}
In the field of information fusion, information fusion methods can achieve better results than a single information source as long as there are appropriate fusion criteria. The traditional pixel-level saliency map fusion method generates a fused image through the weighted sum of multiple saliency maps, which can be expressed as:
\begin{equation}\label{equ:15}
S_{fused}\left( x,y \right)=\sum_{n=1}^{N}W_{n}\left( x,y \right)S_{n}\left( x,y \right),
\end{equation}
where $ N $ is the number of saliency maps to be fused, $ S_{n}\left( x,y \right) $ is the pixel intensity of the $ n $-th saliency map at $ \left( x,y \right) $, and $ W_{n}\left( x,y \right) $ is the weight of the importance of pixel $ S_{n}\left( x,y \right) $ at $ \left( x,y \right) $. Therefore, the key to fusion is designing a reasonable weight.

\begin{figure}[thp!]
	\centering
	\begin{overpic}[width=0.72\columnwidth]{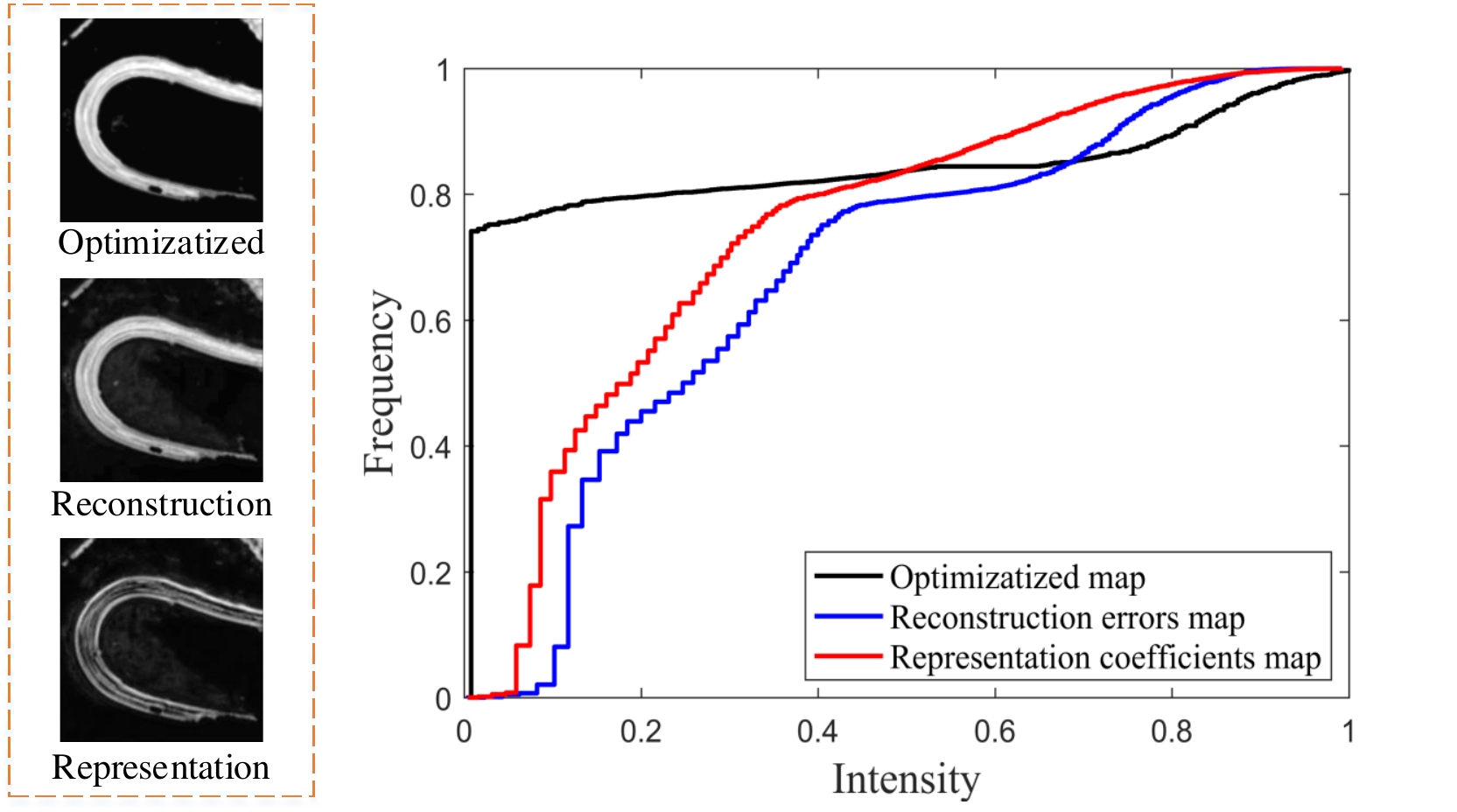}
	\end{overpic}
	\caption{
		Cumulative histogram comparison of different saliency maps.
	}\label{fig:cumulativeHistogram}
	\end{figure}

With this in mind, and based on the observation of the cumulative histogram of pixel intensity (the histogram integral along the pixel intensity axis as shown in \figref{fig:cumulativeHistogram}), we propose a weight function to suppress the background region and highlight the foreground region in the fusion of saliency maps.

\figref{fig:cumulativeHistogram} shows an example of a cumulative histogram of a coefficient representation map, a reconstruction error map to be fused, and an optimized fusion map. The cumulative histogram in the optimized fusion map increases sharply at the beginning; in other words, there are a significant number of pixels in this interval, and the intensity of the surrounding pixels has a small change with a larger gradient than that of the saliency image to be fused. In the middle region of pixel intensity (0.2 to 0.8), the cumulative histogram changes slowly, indicating that there are relatively few pixels in this region, and the surrounding pixel intensity has a greater change that has a smaller gradient compared with the saliency maps to be fused. The analysis for the interval where the pixel intensity is close to 1 follows the same rule as above. Therefore, when the pixels are in the range of a cumulative histogram with a large gradient, they need to be given a higher weight during image fusion. 
Formally, we can express this observation as:
\begin{equation}\label{equ:16}
W_{n}\left( x,y \right)=\frac{Grad_{n}\left( I_{n}\left(x,y\right)\right)}{\sum_{n=1}^{N}Grad_{n}\left( I_{n}\left(x,y\right)\right)+\varphi},
\end{equation}
where $ \varphi $ prevents the occurrence of a very small positive number with zero denominators, and $ Grad_{n}\left( I_{n}\left(x,y\right)\right) $ is the gradient of the cumulative histogram at pixel intensity $ I_{n}\left(x,y\right) $. 
Because the cumulative histogram is the statistical information of all pixels, the gradient in \eqnref{equ:16} is not the local gradient around the pixel; we call it the global gradient. Using the weights obtained above, we can fuse several saliency maps obtained by representation coefficients and reconstruction errors according to \eqnref{equ:15}.
\section{Experiments}
{\color{myred}In this section, we first introduce the constructed RS image dataset and three other popular datasets containing natural images, and then explain the dictionary training strategy, evaluation metrics and implementation details. Finally, we compare the proposed method with nine state-of-the-art methods.}
\subsection{Experimental setup}
\subsubsection{Datasets}
\begin{figure}[thp!]
	\centering
	\begin{overpic}[width=1\columnwidth]{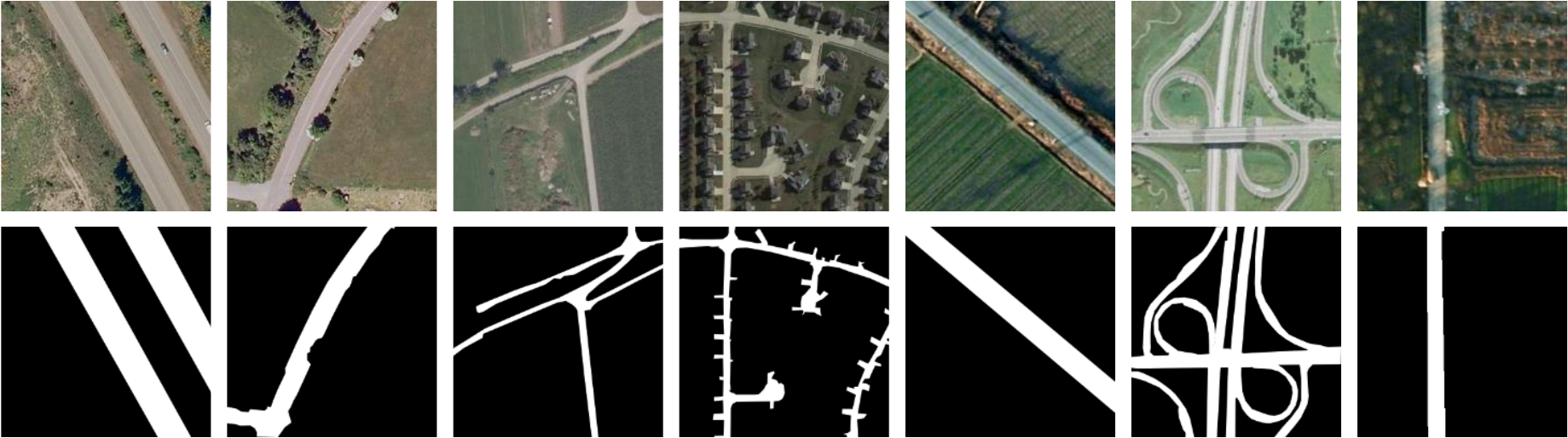}
	\end{overpic}
	\caption{
		Sample images from the constructed 2RSOD dataset. The first row shows the optical RS images. The second row provides the pixel-wise annotations.
	}\label{fig:ExampleImage}
\end{figure}

To the best of our knowledge, no publicly available dataset of optical RS images can be used for road detection. Therefore, we collected 300 optical RS images to build a dataset for road saliency detection, which we called ``2RSOD", and manually annotated each image, pixel-wise. Most of the original optical RS images were collected from Google Earth, and the rest were collected from existing optical RS image datasets, including DOTA \cite{xia2018dota} and NWPU VHR-10  \cite{cheng2016learning}. This 2RSOD dataset is challenging because the spatial resolutions of the images are diverse, including $ 300\times 300 $ , $ 500\times 500 $ and $ 1024\times 1024 $.  Further, image backgrounds tend to be complicated and cluttered, often including buildings, trees, rivers, and shadows. The sizes, numbers, and shapes of the salient objects also vary. Some sample images from the constructed 2RSOD dataset are shown in \figref{fig:ExampleImage}. {\color{myred}In addition, we evaluate CDL on three other benchmark natural image datasets, including ECSSD \cite{yan2013hierarchical} with 1000 images, PASCAL-S \cite{li2014secrets} with 850 images, and DUT-OMRON \cite{yang2013saliency} with 5168 images.}
\subsubsection{Evaluation metrics}
To quantitatively evaluate the performance of various methods, we adopt six evaluation metrics. \tabref{tab:2} summarizes these metrics.
\begin{table}[t!]
	\centering
	\scriptsize
	\caption{Summary of evaluation metrics}\label{tab:2}
	\begin{tabular*}{0.95\linewidth}{ll}
		
		\toprule
		\textbf{Metric} & \textbf{Mathematical Expression} \\
		\midrule
		Precision-Recall $ \left(PR\right) \uparrow $ & $ Precision\left( P\right):\frac{\left|S \cap G \right|}{\left|S \right|} $, $ Recll\left( R\right): \frac{\left|S \cap G \right|}{\left|G \right|} $ \\
		F-measure$ \left(F_{\beta}\right) \uparrow $ & $ F_{\beta}=\left(1+\beta ^2\right)\frac{P*R}{\beta ^{2}P+R},  \beta ^2=0.3 $ \\
		S-measure$ \left(S_{\alpha}\right) \uparrow $ & $ S=\alpha * S_{0}+\left(1-\alpha\right)*S_{r},  \alpha=0.5$\\
		E-measure$ \left(E_{\xi}\right)\uparrow $ &  $ E=\frac{1}{W*H}\sum_{i=1}^{W}\sum_{i=1}^{H}\phi_{FM}\left(i,j\right) $ \\
		Adaptive threshold $ \left(F_{adp}\right)\uparrow $ &  $ Thr=\frac{2}{W*H}\sum_{i=1}^{W}\sum_{i=1}^{H}S\left(i,j\right) $ \\
		Mean absolute error $ \left(MAE ~\mathcal{M}\right) \downarrow $ & $ MEA=\frac{1}{W*H}\sum_{i=1}^{W}\sum_{i=1}^{H}\left|S\left(i,j\right)-G\left(i,j\right) \right| $\\
		\bottomrule
		\multicolumn{2}{l}{Note: $ \uparrow $ \& $ \downarrow $ denote larger and smaller is better.} \\
		$ S $:saliency image & $ S_{o} $:target perception structure \cite{fan2017structure}\\
		$ G $:corresponding annotation map & $ \phi $:enhanced contrast matrix \cite{fan2018enhanced}\\
		$ \left|\cdot \right| $:calculates the number of nonzero entries & $ W $:width of the image \\
		$ S_{r} $:similarity measurement of the region\cite{fan2017structure} & $ H $:height of the image \\
	\end{tabular*}
\end{table}
\subsubsection{Parameter settings}
\begin{figure}
	\centering
	\subfloat[]{\includegraphics[scale=0.52]{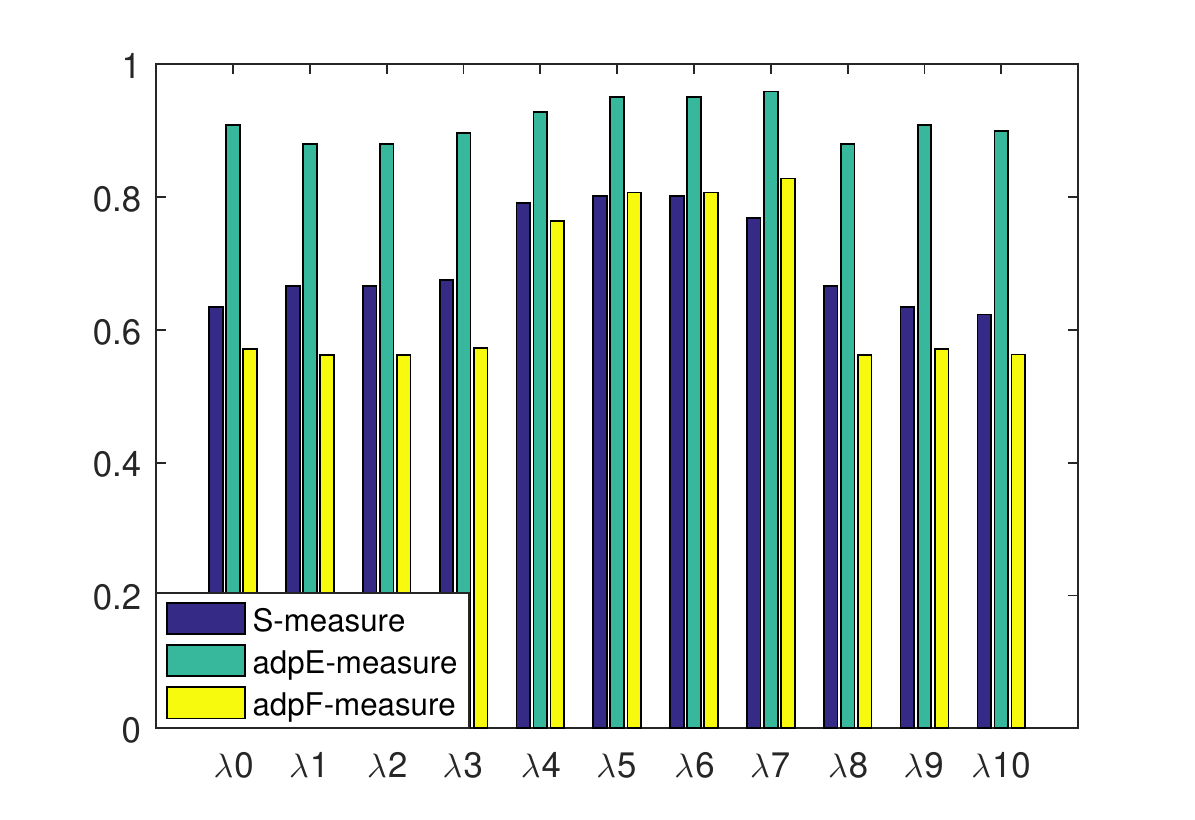}}%
	\subfloat[]{\includegraphics[scale=0.52]{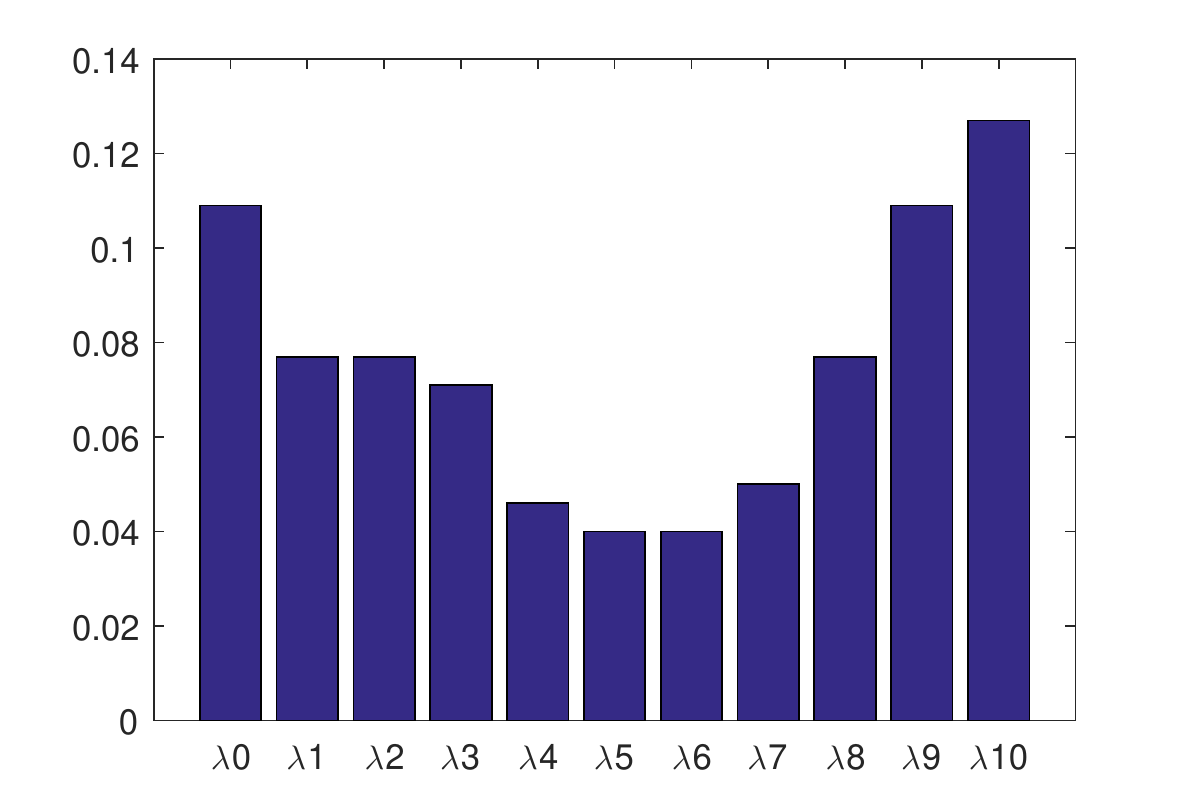}}%
	\caption{Quantitative comparison of various values of $ \lambda_2 $. (a) S-measure, adaptive F-measure and E-measure values. (b) MAE values.}%
	\label{fig:lamda}
\end{figure}

All parameter settings related to our experiment are summarized in \tabref{tab:3}. {\color{myred}For 2RSOD and the other three natural image datasets, we select 240 images from each as the training sets for the discriminant dictionaries, and the remaining images as the test sets.} For the dictionary learning of our proposed CDL algorithm, we sample 480 salient and non-salient image patches of size $ 80\times 80 $  from the training set as training patches. During of dictionary training, following the empirical settings in article \cite{mairal2009online}, we down-sample the training patch into an image patch of size $ 16\times 16 $ as the input, so that the number of pixels $ m $ of the learned dictionary atom is 256, and the number of atoms in the dictionary $ k $ is set to $ 4\times m $ . In \eqnref{equ:5}, the regularization  parameter $ \lambda_1 $ is set to $ 1.2/\sqrt{m} $ to weight the reconstruction error and sparsity, and the learning rate $ \sigma $ is set to 0.02 in dictionary learning (\eqnref{equ:12}) to obtain a more discriminative dictionary. Further, we use various values  $  \lambda0=0.001,\lambda1=0.005,\lambda2=0.01,\lambda3=0.02,\lambda4=0.03,\lambda5=0.04,\lambda6=0.05,\lambda7=0.06,\lambda8=0.07,\lambda9=0.09,\lambda10=0.1 $ for testing with respect to  $\lambda_2$. The experimental results are shown in \figref{fig:lamda}, according to which the parameter $\lambda_2$ in \eqnref{equ:12} is adjusted to 0.05.
\begin{table}[t!]
	\centering
	\scriptsize
	\caption{Parameter settings in our method}\label{tab:3}
	\begin{tabular*}{0.75\linewidth}{lll}
		
		\toprule
		\textbf{Process} & \textbf{Parameter Description} & \textbf{Value} \\
		\midrule
		& Training patch size        & $ 16 \times 16 $ \\
		& Dictionary atom size $ m $ & $ 256 \times 1 $ \\
		Dictionary learning & Number of atoms in the dictionary $ k $& $ 1 \times 1024 $\\
		& Regularisation  parameter $ \lambda_1 $ & 0.02 \\
		& Regularisation  parameter $ \lambda_2 $& 0.02 \\
		\midrule
		& Scalar parameter $ \eta_{\HZupbf{A}} $ & 1 \\
		Saliency detection	& Scalar parameter $ \eta_{\HZupbf{R}} $ & 1\\
		& Positive value $ \varphi $& 0.001\\
		\bottomrule	
	\end{tabular*}
\end{table}
\subsection{Verification and analysis}
In this subsection, we first use the 2RSOD dataset to demonstrate the effectiveness of the CDL-based model, then compare our proposed model with state-of-the-art methods on four datasets, and finally analyze some failure cases in our method.
\subsubsection{Effectiveness analysis based on CDL model}
In this subsection, we analyze and verify the effectiveness of the proposed CDL-based saliency detection model on the 2RSOD dataset from the following four aspects:
\begin{figure}
	\centering
	\subfloat[]{\includegraphics[scale=0.52]{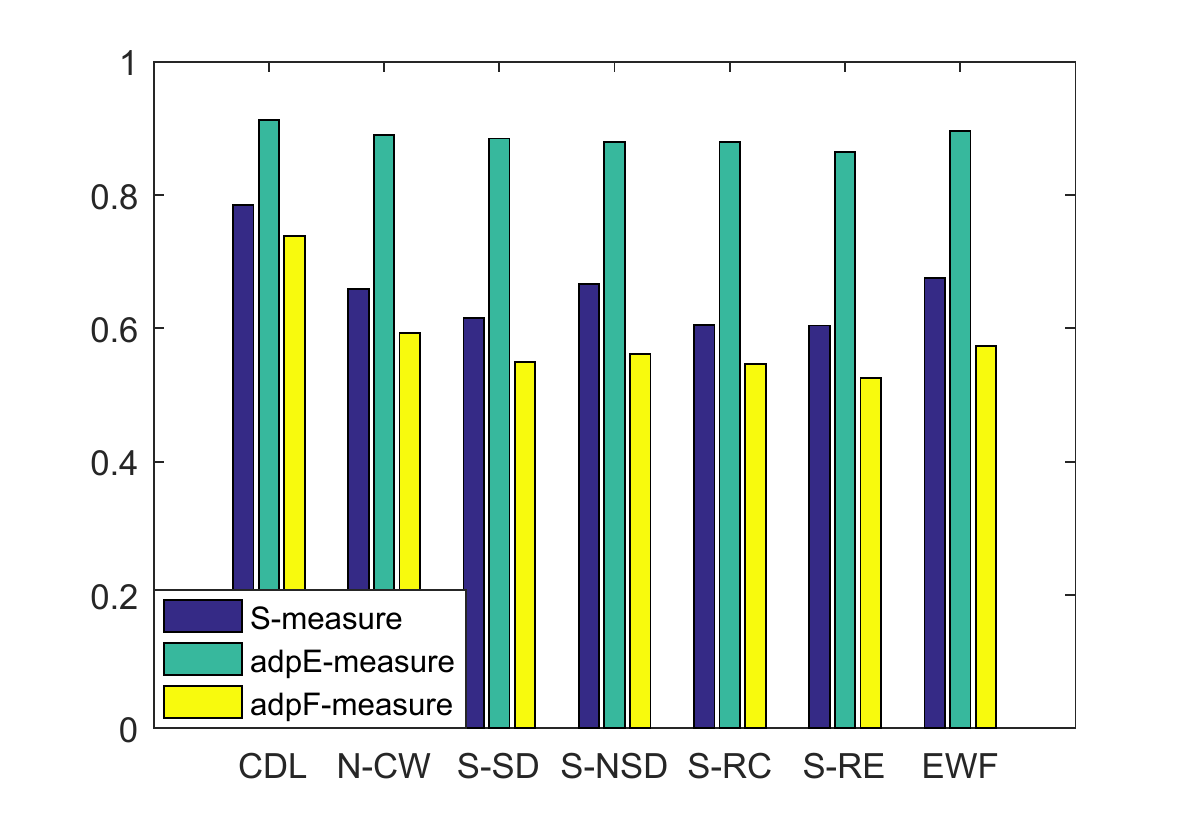}}%
	\subfloat[]{\includegraphics[scale=0.52]{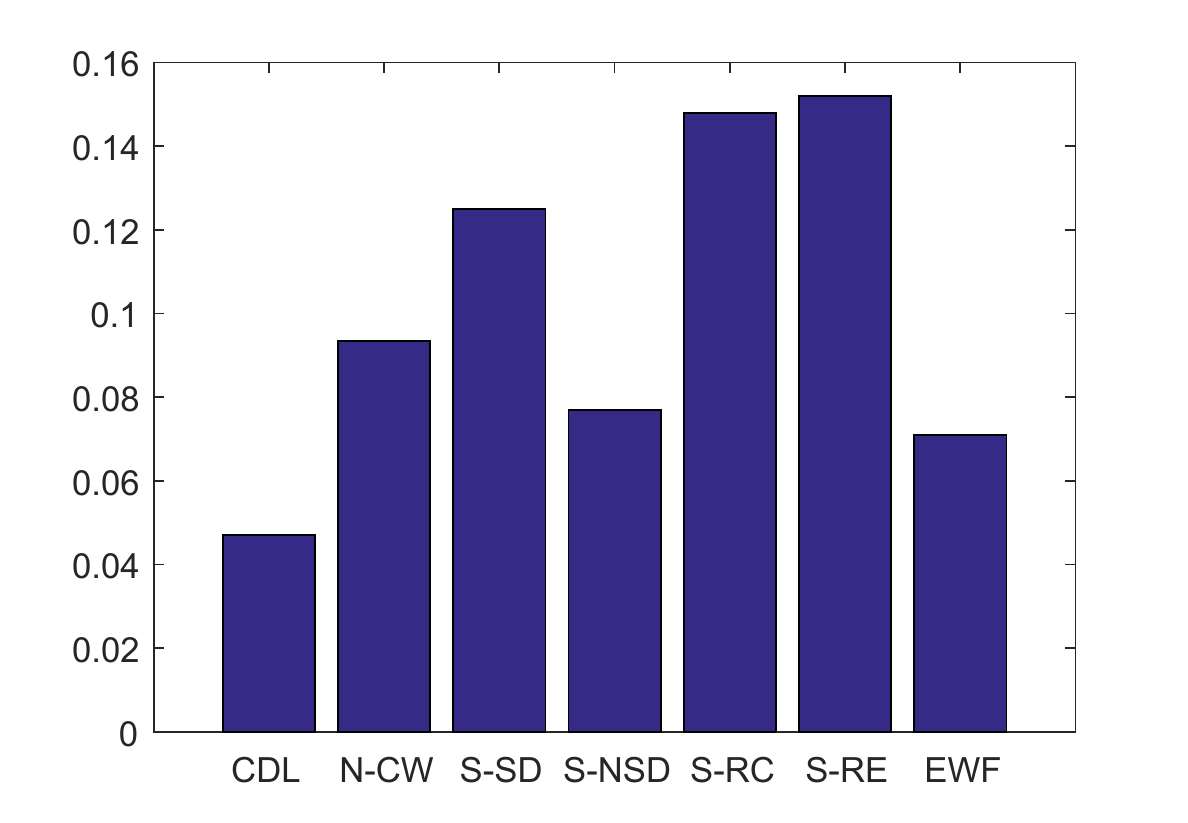}}%
	\caption{Quantitative comparison of various verification methods. (a) S-measure, adaptive F-measure and E-measure values. (b) MAE values.}%
	\label{fig:effectiveness}
\end{figure}

\textbf{A. Effectiveness of contrast-weighted terms}

{\color{myred}The contrast-weighted term in \eqnref{equ:7} is used to optimize the atoms update during the learning process of the discriminant dictionary. We set it to 1 to verify its effectiveness, as shown in N-CW in \figref{fig:effectiveness} (a).}

\textbf{B. Effectiveness of constructing a discriminant dictionary}

Discriminant dictionaries have their own features for the SR of images. To verify the effectiveness of the discriminant dictionary for saliency detection, we use the single salient dictionary (S-SD) or single non-salient dictionary (S-NSD) for saliency detection.

\textbf{C. The validity of significance measures for joint representation coefficients and reconstruction errors}

To improve the expression of outliers in the coefficients of the SR, we combine representation coefficients and reconstruction errors as a measure of saliency detection. To verify the effectiveness of the joint saliency measure, we use the single representation coefficient (S-RC) and the single reconstruction error (S-RE) as the saliency measurements.

\textbf{D. Effectiveness of saliency map fusion method based on global gradient optimization}

To improve the use of correct information in multiple saliency maps, we proposed a saliency map fusion method based on global gradient optimization. We compare the saliency of this proposed optimization method with the equal weight fusion (EWF) method as a verification of the effectiveness of our method.

\figref{fig:effectiveness} shows that the proposed saliency detection method based on CDL is superior to the above effectiveness verification methods, in terms of several evaluation metrics of. The figure also shows the importance and contribution of the various parts that make up the proposed method.

\subsubsection{Comparison with state-of-the-art methods}
{\color{myred}We compare the proposed algorithm with nine state-of-the-art SOD methods, including three traditional methods (LPS \cite{li2015inner} DSG \cite{zhou2017salient}, WMR \cite{zhu2018saliency}), three methods related to SR (SMD \cite{peng2016salient}, RSR-LC \cite{yi2018salient}, RDR \cite{xiao2018salient}), and three latest deep learning-based methods (BMPM \cite{zhang2018bi}, BASNet \cite{qin2019basnet}, PAGE \cite{wang2019salient}). All results are either generated by the source code or provided by the author.}

\textbf{A. Visual comparison}

\begin{figure}[thp!]
	\centering
	\begin{overpic}[width=.99\columnwidth]{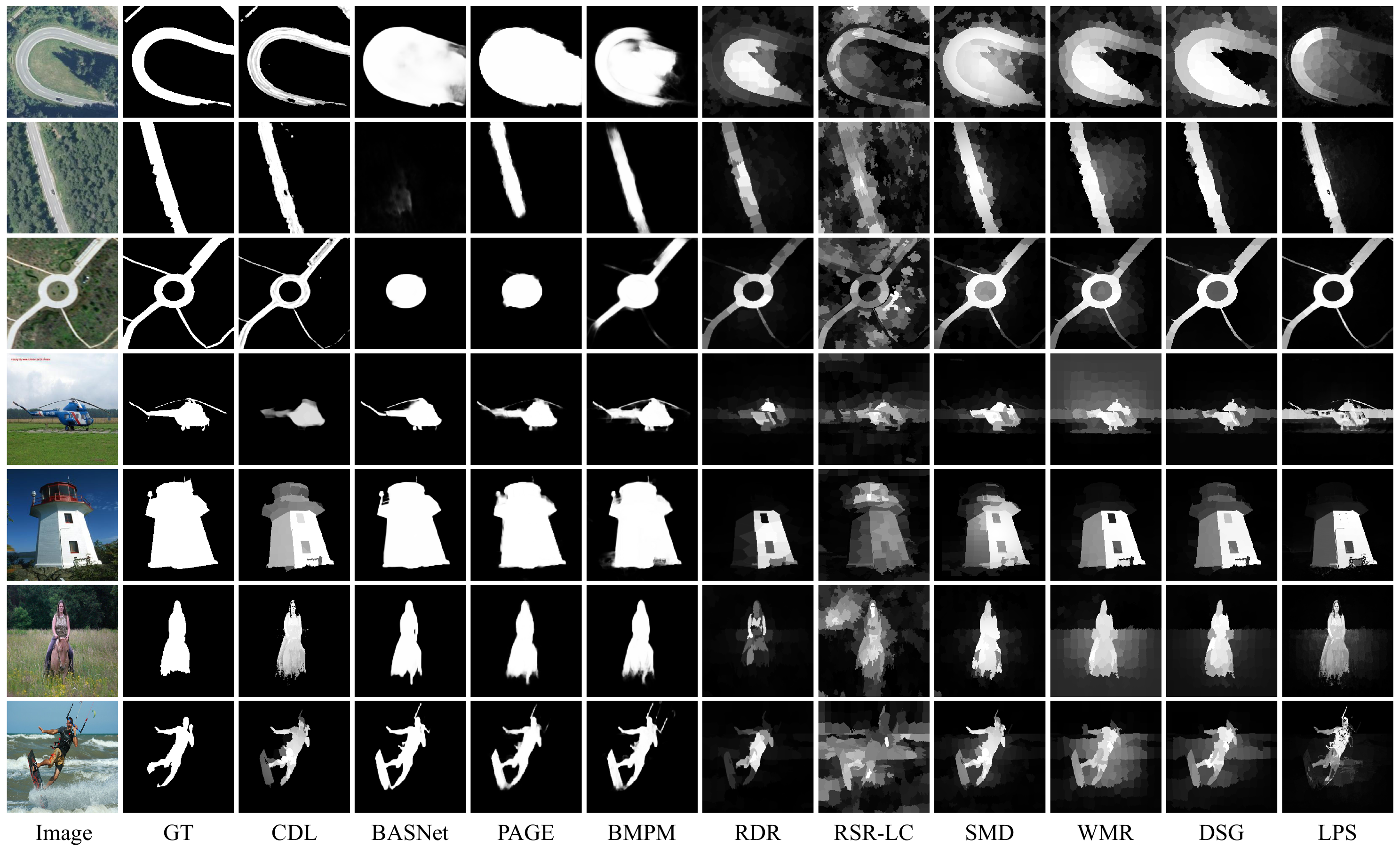}
	\end{overpic}
	\caption{Visual comparisons of various methods.}
	\label{fig:visualCompare}
\end{figure}

{\color{myred}As shown in \figref{fig:visualCompare}, most of the comparison methods perform poorly on 2RSOD. On the other hand, the proposed method is competitive on three natural image datasets. Further observation shows that, for the images with simple backgrounds and prominent foregrounds (for example: the second and sixth rows in \figref{fig:visualCompare}), all methods have better detection results. However, for images with complex backgrounds (for example: the first row in \figref{fig:visualCompare}) that contain shadows, buildings, and so on, the comparison methods do not have satisfactory results. Moreover, because most of the salient road regions of the images in the 2RSOD dataset are linked to image boundaries, the accuracies of saliency detection methods based on boundary priors (for example, WMR \cite{zhu2018saliency}, SMD \cite{peng2016salient}, RSR-LC \cite{yi2018salient}, RDR \cite{xiao2018salient}) were also affected. In contrast, the proposed method can effectively separate the salient object from the background and obtains good detection results for images with complex scenes or similar foreground and background.}

\textbf{B. Quantitative comparison}

\begin{figure}[thp!]
	\centering
	\begin{overpic}[width=.99\columnwidth]{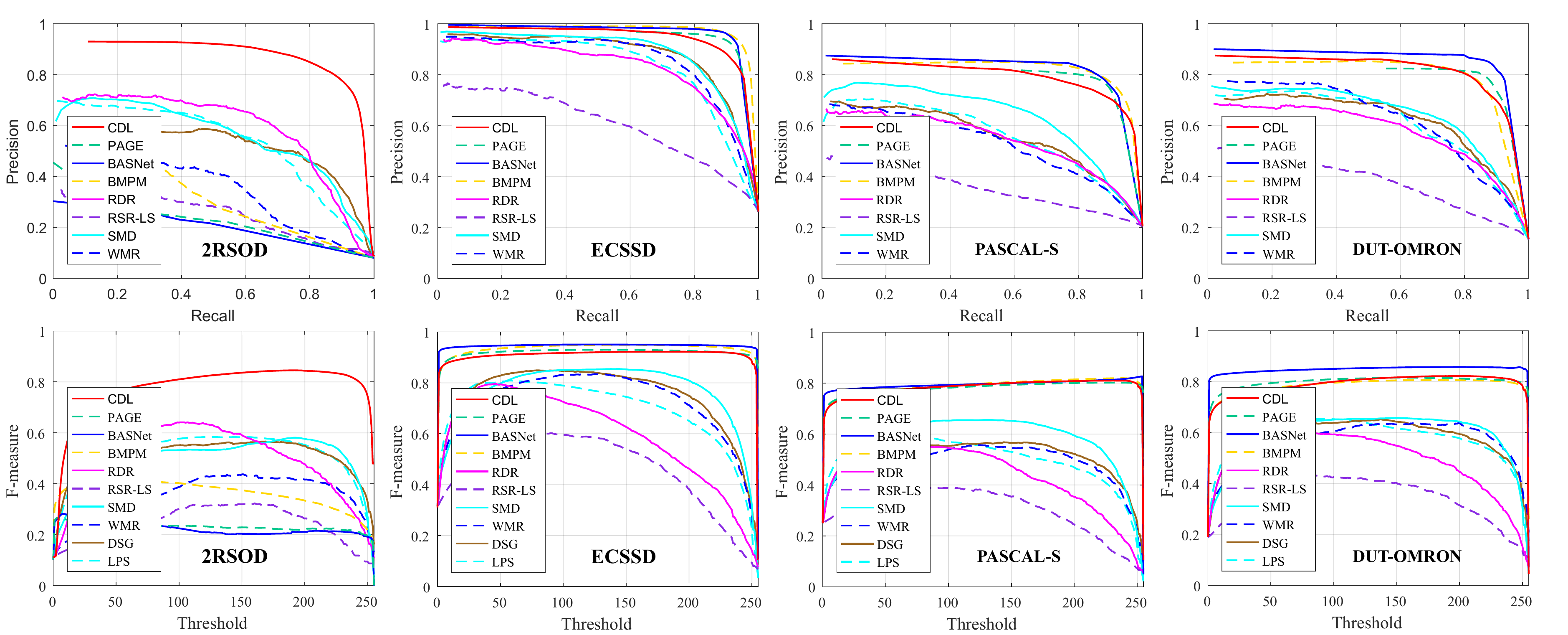}
	\end{overpic}
	\caption{Performance comparison with nine state-of-the-art methods over four datasets. The first row shows a comparison of precision-recall curves. The second row shows a comparison of F-measure curves over different thresholds.}
	\label{fig:compareWithOther}
\end{figure}

{\color{myred}To fully compare the proposed method with the above models, the detailed experimental results in terms of four metrics are listed in \tabref{tab:4}. In addition, \figref{fig:compareWithOther} shows the standard PR curves and the F-measure curves on the four datasets, which can be used to evaluate the holistic performance of models. As shown in the above experimental results, our proposed method is highly competitive under all six metrics, especially on the 2RSOD dataset. At the same time, the method proposed in this paper is significantly better than the saliency detection method related to SR.}
\begin{table}[t!]
	\centering
	\scriptsize
	\renewcommand{\arraystretch}{1.1} 
	\renewcommand{\tabcolsep}{0.68mm} 
	
	\caption{Quantitative evaluation. The mean F-measure, S-measure and MAE of different saliency detection methods on 2RSOD and three benchmark datasets. The best four results are highlighted in {\color{red}{red}}, {\color{blue}{blue}}, {\color{green}{green}} and {\color{violet}{purple}}. $ \dag $ \& $ \ddag $ denote methods based on SR and deep learning. ``PAS-S'' \& ``DUT-O'' represent datasets PASCAL-S and DUT-OMRON.}\label{tab:4}
	\begin{tabular*}{1\linewidth}{clr|ccc|ccc|ccc|c} 
		\toprule
		& & & ~LPS~ & ~DSG~ & ~WMR~~ & ~SMD$^{\dag}$~ & SRS-LC$^{\dag}$ & ~RDR$^{\dag}$ & BMPM$^{\ddag}$ & PAGE$^{\ddag}$& BASNet$^{\ddag}$& CDL \\
		& & Metric &  \cite{li2015inner} & \cite{zhou2017salient} & \cite{zhu2018saliency} & \cite{peng2016salient} & \cite{yi2018salient} & \cite{xiao2018salient} & \cite{zhang2018bi} & \cite{wang2019salient} & \cite{wang2019salient} & (ours)\\
		\midrule
		\multirow{4}{*}{\rotatebox{90}{2RSOD}}&\multirow{4}{0.1cm}{} & $\mathcal{M}\downarrow $& {\color{blue}{0.114}}  & 0.146  & 0.237  & 0.154  & 0.229  & {\color{green}{0.123}}  & 0.172  & 0.169  & {\color{violet}{0.128}}  & {\color{red}{0.063}}  \\
		& & $ F_{\beta}\uparrow $ & {\color{blue}{0.493}}  & {\color{green}{0.492}}  & 0.326  & 0.468  & 0.224  & 0.474  & 0.442  & 0.403  & {\color{violet}{0.485}}  & {\color{red}{0.794}}  \\
		& & $ S_{\alpha}\uparrow $ & {\color{blue}{0.660}}  & {\color{violet}{0.641}}  & 0.530  & 0.633  & 0.503  & {\color{green}{0.646}}  & 0.607  & 0.587  & 0.628  & {\color{red}{0.813}}  \\
		& & $ E_{\xi}\uparrow $ & {\color{green}{0.811}} & 0.709  & 0.629  & {\color{violet}{0.772}}  & 0.729  & {\color{blue}{0.818}}  & 0.734  & 0.701  & 0.743  & {\color{red}{0.901}}  \\
		\midrule
		\multirow{4}{*}{\rotatebox{90}{ECSSD}}&\multirow{4}{0.1cm}{\rotatebox{90}{\cite{yan2013hierarchical}}} & $\mathcal{M}\downarrow $& 0.169  & 0.146  & 0.162  & 0.141  & 0.257  & 0.198  & {\color{green}{0.044}}  & {\color{blue}{0.042}}  & {\color{red}{0.037}}  & {\color{violet}{0.061}}  \\
		& & $ F_{\beta}\uparrow $ & 0.629  & 0.701  & 0.669  & 0.725  & 0.422  & 0.549  & {\color{blue}{0.894}}  & {\color{red}{0.906}}  & {\color{green}{0.880}}  & {\color{violet}{0.879}}  \\
		& & $ S_{\alpha}\uparrow $ & 0.700  & 0.773  & 0.754  & 0.795  & 0.592  & 0.641  & {\color{green}{0.911}}  & {\color{blue}{0.912}}  & {\color{red}{0.916}}  & {\color{violet}{0.884}}  \\
		& & $ E_{\xi}\uparrow $ & 0.768  & 0.823  & 0.820  & 0.839  & 0.702  & 0.756  & {\color{green}{0.914}}  & {\color{blue}{0.920}}  & {\color{red}{0.921}}  & {\color{violet}{0.898}}  \\
		\midrule
		\multirow{4}{*}{\rotatebox{90}{PAS-S}}&\multirow{4}{0.1cm}{\rotatebox{90}{\cite{li2014secrets}}} & $\mathcal{M}\downarrow $& 0.203  & 0.231  & 0.249  & 0.198  & 0.276  & 0.218  & {\color{red}{0.037}}  & {\color{green}{0.077}}  & {\color{blue}{0.076 }} & {\color{violet}{0.088}}  \\
		& & $ F_{\beta}\uparrow $ & 0.425  & 0.446  & 0.428  & 0.514  & 0.269  & 0.382  & {\color{blue}{0.803}}  & {\color{red}{0.810}}  & {\color{green}{0.775}}  & {\color{violet}{0.765}}  \\
		& & $ S_{\alpha}\uparrow $ & 0.580  & 0.595  & 0.587  & 0.661  & 0.496  & 0.564  & {\color{red}{0.840}}  & {\color{blue}{0.835}}  & {\color{green}{0.832}}  & {\color{violet}{0.823}}  \\
		& & $ E_{\xi}\uparrow $ & 0.677  & 0.694  & 0.674  & 0.733  & 0.646  & 0.686  & {\color{violet}{0.838}}  & {\color{green}{0.841}}  & {\color{red}{0.847}}  & {\color{blue}{0.844}}  \\
		\midrule
		\multirow{4}{*}{\rotatebox{90}{DUT-O}}&\multirow{4}{0.1cm}{\rotatebox{90}{\cite{yang2013saliency}}} & $\mathcal{M}\downarrow $& 0.135  & 0.180  & 0.197  & 0.160  & 0.230  & 0.161  & {\color{violet}{0.063}}  & {\color{blue}{0.052}}  & {\color{red}{0.048}}  & {\color{green}{0.061}}  \\
		& & $ F_{\beta}\uparrow $ & 0.530  & 0.520  & 0.504  & 0.537  & 0.296  & 0.442  & {\color{violet}{0.698}}  & {\color{blue}{0.777}}  & {\color{red}{0.791}}  & {\color{green}{0.762}}  \\
		& & $ S_{\alpha}\uparrow $ & 0.678  & 0.663  & 0.650  & 0.690  & 0.546  & 0.624  & {\color{violet}{0.809}}  & {\color{blue}{0.854}}  & {\color{red}{0.866}}  & {\color{green}{0.840}}  \\
		& & $ E_{\xi}\uparrow $ & 0.748  & 0.760  & 0.729  & 0.746  & 0.659  & 0.731  & {\color{violet}{0.839}}  & {\color{blue}{0.869}}  & {\color{red}{0.884}}  & {\color{green}{0.860}}  \\
		\bottomrule	
	\end{tabular*}
\end{table}

\textbf{C. Computational complexity comparison}

{\color{myred}To demonstrate the computational efficiency of the proposed method, we test the average execution time of several state-of-the-art methods and the proposed method on the 2RSOD dataset. These methods are run on a desktop with an Intel Core i7-7700 CPU and RTX 2070 GPU. As shown in \tabref{tab:5}, the efficiency of the CDL method based on Matlab programming can reach the average of the comparison method.}
\begin{table}[t!]
	\centering
	\scriptsize
	 \renewcommand{\arraystretch}{1.1}
	\renewcommand{\tabcolsep}{1mm}
	\caption{Average execution time of several methods}\label{tab:5}
	\begin{tabular*}{0.95\linewidth}{c|ccc|ccc|ccc|c}
		\toprule
		\multirow{2}{*}{\textbf{Method}}& LPS & DSG & WMR & SMD$^{\dag}$ & SRS-LC$^{\dag}$  & RDR$^{\dag}$  & BMPM$^{\ddag}$  & PAGE$^{\ddag}$  & BASNet$^{\ddag}$  & CDL \\
		  &  \cite{li2015inner} & \cite{zhou2017salient} & \cite{zhu2018saliency} & \cite{peng2016salient} & \cite{yi2018salient} & \cite{xiao2018salient} & \cite{zhang2018bi} & \cite{wang2019salient} & \cite{wang2019salient} & (ours) \\
		\midrule
		\textbf{Time(s)} & 2.21 & 0.83 & 1.56 & 0.91 & 3.32 & 2.48 & 1.26 & 0.12 & 0.68 & 1.52 \\
		\bottomrule	
	\end{tabular*}
\end{table}
\subsubsection{Failure cases}
\begin{figure}[thp!]
	\centering
	\begin{overpic}[width=.98\columnwidth]{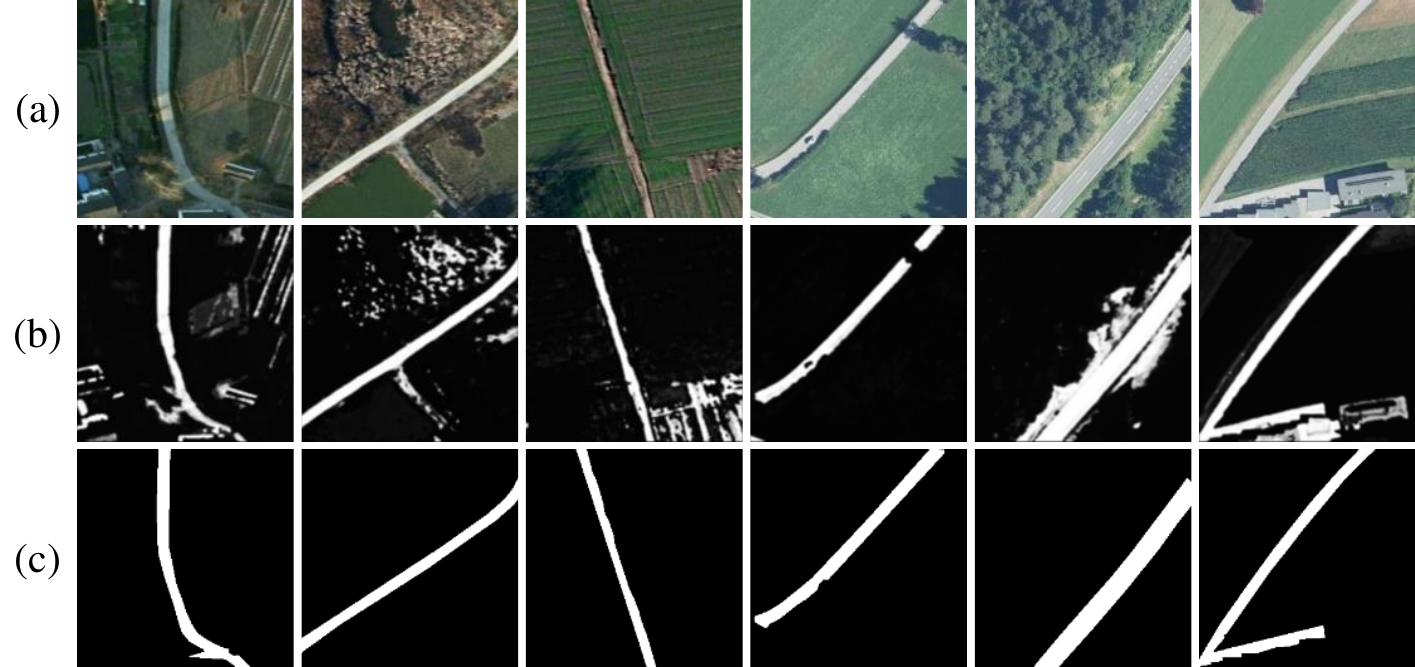}
	\end{overpic}
	\caption{Failure cases of proposed method. (a) Original images. (b) Saliency maps obtained by the proposed method. (c) Ground truth.}.
	\label{fig:failureCases}
\end{figure}

Although the proposed method can accurately detect most salient road  regions, there are still some limitations.
 \figref{fig:failureCases} shows that when an image contains regions with similar appearance to the road (such as roofs, or farmland), our proposed method incorrectly marks the background regions as the foreground. 
 Also, the places where the road regions would be interrupted are shown in the third column of \figref{fig:failureCases}, which is inconsistent with the fact that the road has connectivity.
 On the other hand, the saliency of the road should be regional and overall, but as shown in the first and second columns of \figref{fig:failureCases}, there are many scattered points with high saliency values in the inspection results. {\color{myred}Through the above analysis, we can construct a more robust dictionary to overcome these problems by combining the semantic information of the image (similar to the work of \cite{liu2018semantic} and \cite{li2017multi}) and the feature information of the target, which is one of our future works.}
\section{Conclusion}
{\color{myred}In this paper, we propose a novel saliency detection method for RS images based on SR. According to the characteristics of salient and non-salient regions, our method uses the proposed online discriminant dictionary learning algorithm to introduce contrast-weighted items into the dictionary learning process to construct a discriminant dictionary based on optimized contrast weighted atoms. Under the discriminant dictionary, we combine the representation coefficients and reconstruction errors of image blocks as saliency detection metrics to generate multiple saliency maps. Considering the complementary information between saliency maps, we propose a saliency map fusion method based on global gradient optimization to integrate multiple saliency maps, which further improves the use of important information from these saliency maps. In addition, we collected and annotated a dataset containing 300 optical RS images. Qualitative, quantitative and ablation experiments on this dataset verify the effectiveness of the proposed method. However, we find that the detection method may fail if an image contains high-contrast areas or has areas similar to the foreground, and the efficiency of the algorithm needs to be improved.
	
In future work, we will combine the semantic information of the scene and the feature information of the salient object, and develop a more accurate color dictionary to improve the robustness of the multi-class saliency detection. Further, due to the successful use of depth information in SOD, we will explore its application in stereo paired RS data. Inspired by recent work \cite{fan2019scoot} and considering the large-scale and final processing structure of RS images, we also plan to introduce the structure co-occurrence texture (scoot) as a perceptual metric for future SOD work.
}

\section*{Acknowledgments}
This research was supported by a grant from the Sichuan Major Science and Technology Special Foundation (No.2018GZDZX0017).
	
	
	

%
%
%
    \bibliographystyle{elsarticle-num} 
	\bibliography{LCWA.bib}
\end{document}